\pdfoutput=1
\documentclass[11pt,a4paper]{article}
\usepackage[hyperref]{sty}
\usepackage{times}
\usepackage{latexsym}

\usepackage{enumitem}
\setitemize{noitemsep,topsep=0pt,parsep=0pt,partopsep=0pt}
\setenumerate{noitemsep,topsep=0pt,parsep=0pt,partopsep=0pt}

\usepackage{graphicx}
\graphicspath{ {./fig/} }

\usepackage{stfloats}
\usepackage{booktabs}
\usepackage{fixltx2e}

\usepackage{microtype}

\aclfinalcopy 

\usepackage{xcolor}
\definecolor{todo}{rgb}{1,0.5,0}

\title{A Matter of Framing:\\ The Impact of Linguistic Formalism on Probing Results}

\author{Ilia Kuznetsov \and Iryna Gurevych \\
  Ubiquitous Knowledge Processing Lab (UKP-TUDA) \\
  Department of Computer Science, Technische Universit\"at Darmstadt \\
  {\texttt http://www.ukp.tu-darmstadt.de/}}
\date{}

\date{}

\begin{document}
\maketitle
\begin{abstract}
Deep pre-trained contextualized encoders like BERT \cite{bert} demonstrate remarkable performance on a range of downstream tasks. A recent line of research in probing investigates the linguistic knowledge implicitly learned by these models during pre-training. While most work in probing operates on the task level, linguistic tasks are rarely uniform and can be represented in a variety of formalisms. Any linguistics-based probing study thereby inevitably commits to the formalism used to annotate the underlying data. Can the choice of formalism affect probing results? To investigate, we conduct an in-depth cross-formalism layer probing study in role semantics. We find linguistically meaningful differences in the encoding of semantic role- and proto-role information by BERT depending on the formalism and demonstrate that layer probing can detect subtle differences between the implementations of the same linguistic formalism. Our results suggest that linguistic formalism is an important dimension in probing studies, along with the commonly used cross-task and cross-lingual experimental settings.
\end{abstract}

\section{Introduction}

The emergence of deep pre-trained contextualized encoders has had a major impact on the field of natural language processing. Boosted by the availability of general-purpose frameworks like \mbox{AllenNLP} \cite{allennlp} and \mbox{Transformers} \cite{transformers}, pre-trained models like ELMO \cite{elmo} and BERT \cite{bert} have caused a shift towards simple architectures where a strong pre-trained encoder is paired with a shallow downstream model, often outperforming the intricate task-specific architectures of the past.

\begin{figure}
	\includegraphics[width=8cm]{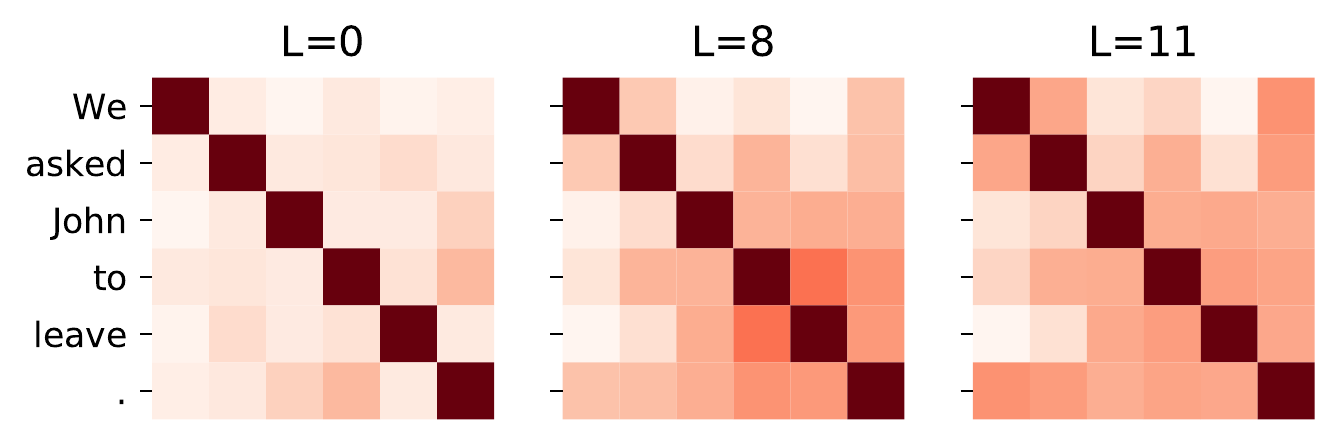}
	\caption{Intra-sentence similarity by layer $L$ of the multilingual BERT-base. Functional words are similar in $L=0$, syntactic groups emerge at higher levels.}
	\label{fig:intersim}
\end{figure}

The versatility of pre-trained representations implies that they encode some aspects of general linguistic knowledge \cite{geometry}. Indeed, even an informal inspection of layer-wise intra-sentence similarities (Fig. \ref{fig:intersim}) suggests that these models capture elements of linguistic structure, and those differ depending on the layer of the model. A grounded investigation of these regularities allows to interpret the model's behavior, design better pre-trained encoders and inform the downstream model development. Such investigation is the main subject of probing, and recent studies confirm that BERT implicitly captures many aspects of language use, lexical semantics and grammar \cite{bertology}.

Most probing studies use linguistics as a theoretical scaffolding and operate on a task level. However, there often exist multiple ways to represent the same linguistic task: for example, English dependency syntax can be encoded using a variety of \textit{formalisms}, incl. Universal \cite{udep}, Stanford \cite{sdep} and CoNLL-2009 dependencies \cite{conll2009}, all using different label sets and syntactic head attachment rules. Any probing study inevitably commits to the specific theoretical framework used to produce the underlying data. The differences between linguistic formalisms, however, can be substantial. 

Can these differences affect the probing results? This question is intriguing for several reasons. Linguistic formalisms are well-documented, and if the choice of formalism indeed has an effect on probing, cross-formalism comparison will yield new insights into the linguistic knowledge obtained by contextualized encoders during pre-training. If, alternatively, the probing results remain stable despite substantial differences between formalisms, this prompts a further scrutiny of what the pre-trained encoders in fact encode. Finally, on the reverse side, cross-formalism probing might be used as a tool to empirically compare the formalisms and their language-specific implementations. To the best of our knowledge we are the first to explicitly address the influence of formalism on probing.

Ideally, the task chosen for a cross-formalism study should be encoded in multiple formalisms using the same textual data to rule out the influence of the domain and text type. While many linguistic corpora contain several layers of linguistic information, having the same textual data annotated with multiple formalisms for the \textit{same} task is rare. We focus on role semantics -- a family of shallow semantic formalisms at the interface between syntax and propositional semantics that assign roles to the participants of natural language utterances, determining \textit{who} did \textit{what} to \textit{whom}, \textit{where}, \textit{when} etc. Decades of research in theoretical linguistics have produced a range of role-semantic frameworks that have been operationalized in NLP: syntax-driven PropBank \cite{propbank}, coarse-grained VerbNet \cite{verbnet}, fine-grained FrameNet \cite{framenet}, and, recently, decompositional Semantic Proto-Roles (SPR) \cite{protoroles, protoroles2}. The SemLink project \cite{semlink} offers parallel annotation for PropBank, VerbNet and FrameNet for English. This allows us to isolate the object of our study: apart from the role-semantic labels, the underlying data and conditions for the three formalisms are identical. SR3DE \cite{sr3de} provides compatible annotation in three formalisms for German, enabling cross-lingual validation of our results. Combined, these factors make role semantics an ideal target for a cross-formalism probing study.

A solid body of evidence suggests that encoders like BERT capture syntactic and lexical-semantic properties, but only few studies have considered probing for predicate-level semantics \cite{edgeprobing, secrets}. To the best of our knowledge we are the first to conduct a cross-formalism probing study on role semantics, thereby contributing to the line of research on how and whether pre-trained BERT encodes higher-level semantic phenomena.  

\paragraph{Contributions.} This work studies the effect of the linguistic formalism on probing results. We conduct cross-formalism experiments on PropBank, VerbNet and FrameNet role prediction in English and German, and show that the \textbf{formalism can affect probing results} in a linguistically meaningful way; in addition, we demonstrate that layer probing can detect subtle differences between implementations of the same formalism in different languages. On the technical side, we advance the recently introduced edge and layer probing framework \cite{edgeprobing}; in particular, we introduce \textbf{anchor tasks} - an analytical tool inspired by feature-based systems that allows deeper qualitative insights into the pre-trained models' behavior. Finally, advancing the current knowledge about the encoding of predicate semantics in BERT, we perform a fine-grained semantic proto-role probing study and demonstrate that \textbf{semantic proto-role properties can be extracted from pre-trained BERT}, contrary to the existing reports. Our results suggest that along with task and language, linguistic formalism is an important dimension to be accounted for in probing research.

\section{Related Work}

\subsection{BERT as Encoder}

BERT is a Transformer \cite{transformer} encoder pre-trained by jointly optimizing two unsupervised objectives: masked language model and next sentence prediction. It uses WordPiece (WP, \citet{wordpiece}) subword tokens along with positional embeddings as input, and gradually constructs sentence representations by applying token-level self-attention pooling over a stack of layers $L$. The result of BERT encoding is a layer-wise representation of the input wordpiece tokens with higher layers representing higher-level abstractions over the input sequence. Thanks to the joint pre-training objective, BERT can encode words and sentences in a unified fashion: the encoding of a sentence or a sentence pair is stored in a special token \textit{[CLS]}.


To facilitate multilingual experiments, we use the multilingual BERT-base (mBERT) published by \citet{bert}. Although several recent encoders have outperformed BERT on benchmarks \cite{roberta, albert, t5}, we use the original BERT architecture, since it allows us to inherit the probing methodology and to build upon the related findings.

\subsection{Probing}

Due to space limitations we omit high-level discussions on benchmarking \cite{glue} and sentence-level probing \cite{conneau2018}, and focus on the recent findings related to the representation of linguistic structure in BERT. Surface-level information generally tends to be represented in the lower layers of deep encoders, while higher layers store hierarchical and semantic information \cite{belinkov2017, lin2019}. \citet{pipeline} show that the abstraction strategy applied by the English pre-trained BERT encoder follows the order of the classical NLP pipeline. Strengthening the claim about linguistic capabilities of BERT, \citet{hewitt2019syn} demonstrate that BERT implicitly learns syntax, and \citet{geometry} show that it encodes fine-grained lexical-semantic distinctions. \citet{bertology} provide a comprehensive overview of BERT's properties discovered to date.

While recent results indicate that BERT successfully represents lexical-semantic and grammatical information, the evidence of its high-level semantic capabilities is inconclusive. \citet{pipeline} show that the English PropBank semantics can be extracted from the encoder and follows syntax in the layer structure. However, out of all formalisms PropBank is most closely tied to syntax, and the results on proto-role and relation probing do not follow the same pattern. \citet{secrets} identify two attention heads in BERT responsible for FrameNet relations. However, they find that disabling them in a fine-tuning evaluation on the GLUE \cite{glue} benchmark does not result in decreased performance. 

Although we are not aware of any systematic studies dedicated to the effect of formalism on probing results, the evidence of such effects is scattered across the related work: for example, the aforementioned results in \citet{pipeline} show a difference in layer utilization between constituents- and dependency-based syntactic probes and semantic role and proto-role probes. It is not clear whether this effect is due to the differences in the underlying datasets and task architecture, or the formalism per se.

Our probing methodology builds upon the edge and layer probing framework. The encoding produced by a frozen BERT model can be seen as a layer-wise snapshot that reflects how the model has constructed the high-level abstractions. \citet{edgeprobing} introduce the edge probing task design: a simple classifier is tasked with predicting a linguistic property given a pair of spans encoded using a frozen pre-trained model. \citet{pipeline} uses edge probing to analyze the layer utilization of a pre-trained BERT model via scalar mixing weights \cite{elmo} learned during training. We revisit this framework in Section \ref{sec:method}.




\subsection{Role Semantics}

We now turn to the object of our investigation: role semantics. For further discussion, consider the following synthetic example:

\begin{enumerate}[label={\alph*.}]
	\item \textit{[John]\textsubscript{Ag} gave [Mary]\textsubscript{Rc} a [book]\textsubscript{Th}.}
	\item \textit{[Mary]\textsubscript{Rc} was given a [book]\textsubscript{Th} by [John]\textsubscript{Ag}.}
\end{enumerate}

Despite surface-level differences, the sentences express the same meaning, suggesting an underlying semantic representation in which these sentences are equivalent. One such representation is offered by role semantics - a shallow predicate-semantic formalism closely related to syntax. In terms of role semantics, \textit{Mary}, \textit{book} and \textit{John} are \textbf{semantic~arguments} of the \textbf{predicate} \textit{give}, and are assigned \textbf{roles} from a pre-defined inventory, for example, \texttt{Agent}, \texttt{Recipient} and \texttt{Theme}.

Semantic roles and their properties have received extensive attention in linguistics \cite{caseforcase, levinhovav2005, dowty} and are considered a universal feature of human language. The size and organization of the role and predicate inventory are subject to debate, giving rise to a variety of role-semantic formalisms.

\textbf{PropBank} assumes a predicate-independent labeling scheme where predicates are distinguished by their sense (\texttt{get.01}), and semantic arguments are labeled with generic numbered core (\texttt{Arg0-5}) and modifier (e.g. \texttt{AM-TMP}) roles. Core roles are not tied to specific definitions, but the effort has been made to keep the role assignments consistent for similar verbs; \texttt{Arg0} and \texttt{Arg1} correspond to the Proto-Agent and Proto-Patient roles as per \citet{dowty}. The semantic interpretation of core roles depends on the predicate sense.

\textbf{VerbNet} follows a different categorization scheme. Motivated by the regularities in verb behavior, \citet{levin} has introduced the grouping of verbs into intersective classes (ILC). This methodology has been adopted by VerbNet: for example, the VerbNet class \texttt{get-13.5.1} would include verbs \textit{earn}, \textit{fetch}, \textit{gain} etc. A verb in VerbNet can belong to several classes corresponding to different senses; each class is associated with a set of roles and licensed syntactic transformations. Unlike PropBank, VerbNet uses a set of approx. 30 thematic roles that have universal definitions and are shared among predicates, e.g. \texttt{Agent}, \texttt{Beneficiary}, \texttt{Instrument}.

\textbf{FrameNet} takes a meaning-driven stance on the role encoding by modeling it i{}n terms of frame semantics: predicates are grouped into frames (e.g. \texttt{Commerce\_buy}), which specify role-like slots to be filled. FrameNet offers fine-grained frame distinctions, and roles in FrameNet are frame-specific, e.g. \texttt{Buyer}, \texttt{Seller} and \texttt{Money}. The resource accompanies each frame with a description of the situation and its core and peripheral participants.

\textbf{SPR} follows the work of \citet{dowty} and discards the notion of categorical semantic roles in favor of feature bundles. Instead of a fixed role label, each argument is assessed via a 11-dimensional cardinal feature set including Proto-Agent and Proto-Patient properties like \texttt{volitional}, \texttt{sentient}, \texttt{destroyed}, etc. The feature-based approach eliminates some of the theoretical issues associated with categorical role inventories and allows for more flexible modeling of role semantics. 

Each of the role labeling formalisms offers certain advantages and disadvantages \cite{fnvnpb, sr3de}. While being close to syntax and thereby easier to predict, PropBank doesn't contribute much semantics to the representation. On the opposite side of the spectrum, FrameNet offers rich predicate-semantic representations for verbs and nouns, but suffers from high granularity and coverage gaps \cite{simpleframeid}. VerbNet takes a middle ground by following grammatical criteria while still encoding coarse-grained semantics, but only focuses on verbs and core (not modifier) roles. SPR avoids the granularity-generalization trade-off of the categorical inventories, but is yet to find its way into practical NLP applications.

\section{Probing Methodology}
\label{sec:method}

We take the edge probing setup by \citet{edgeprobing} as our starting point. Edge probing aims to predict a label given a pair of contextualized span or word encodings. More formally, we encode a WP-tokenized sentence $[wp_1, wp_2, ... wp_k]$ with a frozen pre-trained model, producing contextual embeddings $[e_1, e_2, ... e_k]$, each of which is a layered representation over $L = {\{l_0, l_1, ... l_m\}}$ layers, with encoding at layer $l_n$ for the wordpiece $wp_i$ further denoted as $e^n_i$. A trainable scalar mix is applied to the layered representation to produce the final encoding given the per-layer mixing weights $\{a_0, a_1 .. a_m\}$ and a scaling parameter $\gamma$: $$\overline{e}_i = \gamma\sum_{l=0}^{m}softmax(a^l)e^l_i$$

Given the source $src$ and target $tgt$ wordpieces encoded as $\overline{e}_{src}$ and $\overline{e}_{tgt}$, our goal is to predict the label $y$. 

Due to its task-agnostic architecture, edge probing can be applied to a wide variety of unary (by omitting $tgt$) and binary labeling tasks in a unified manner, facilitating the cross-task comparison. The original setup has several limitations that we address in our implementation.

\textbf{Regression tasks.} The original edge probing only considers classification tasks. Many language phenomena - including positional information and semantic proto-roles, are naturally modeled as regression, and we extend the original model by supporting both classification and regression: the former achieved via softmax, the latter via direct linear regression to the target value.

\textbf{Flat model} To decrease the models' own expressive power \cite{selectivity}, we keep the number of parameters in our probing model as low as possible. While \citet{edgeprobing} utilize pooled self-attentional span representations and a projection layer to enable cross-model comparison, we directly feed the wordpiece encoding into the classifier, using the first wordpiece of a word. To further increase the selectivity of the model, we directly project the source and target wordpiece representations into the label space, opposed to the two-layer MLP classifier used in the original setup.

\textbf{Separate scalar mixes.} To enable fine-grained analysis of probing results, we train and analyze separate scalar mixes for source and target wordpieces, motivated by the fact that the classifier might utilize different aspects of their representation for prediction\footnote{The original work \cite[Appendix C]{edgeprobing}, also uses separate projections for source and target tokens in the background, but does not investigate the differences between the learned projections.}. Indeed, we find that the mixing weights learned for source and target wordpieces might show substantial -- and linguistically meaningful -- variation.

\textbf{Sentence-level probes.} Utilizing the BERT-specific sentence representation \textit{[CLS]} allows us to incorporate the sentence-level natural language inference (NLI) probe into our kit.

\paragraph{Anchor tasks} We employ two analytical tools from the original layer probing setup. Mixing weight plotting compares layer utilization among tasks by visually aligning the respective learned weight distributions transformed via a softmax function. Layer center-of-gravity is used as a summary statistic for a task's layer utilization.

While the distribution of mixing weights along the layers allows us to estimate the order in which information is processed during encoding, it doesn't allow to directly assess the \emph{similarity} between the layer utilization of the probing tasks. 

\citet{pipeline} have demonstrated that the order in which linguistic information is stored in BERT mirrors the traditional NLP pipeline. A prominent property of the NLP pipelines is their use of low-level features to predict downstream phenomena. In the context of layer probing, probing tasks can be seen as end-to-end feature extractors. Following this intuition, we define two groups of probing tasks: \emph{target tasks} -- the main tasks under investigation, and \emph{anchor tasks} -- a set of related tasks that serve as a basis for qualitative comparison between the targets. The softmax transformation of the scalar mixing weights allows to treat them as probability distributions: the higher the mixing weight of a layer, the more likely the probe is to utilize information from this layer during prediction. We use Kullback-Leibler divergence to compare target tasks (e.g. role labeling in different formalisms) in terms of their similarity to lower-level anchor tasks (e.g. dependency relation and lemma). Note that the notion of anchor task is contextual: the same task can serve as a target and as an anchor, depending on the focus of the study.

\section{Setup}

\subsection{Source data}


For German we use the SR3de corpus \cite{sr3de} that contains parallel PropBank, FrameNet and VerbNet annotations for verbal predicates. For English, \mbox{SemLink} \cite{semlink} provides mappings from the original PropBank corpus annotations to the corresponding FrameNet and VerbNet senses and semantic roles. We use these mappings to enrich the CoNLL-2009 \cite{conll2009} dependency role labeling data -- also based on the original PropBank -- with roles in all three formalisms via a semi-automatic token alignment procedure. 
The resulting corpus is substantially smaller than the original, but still an order of magnitude larger than SR3de (Table \ref{tab:datasets}). Both corpora are richly annotated with linguistic phenomena on word level, including part-of-speech, lemma and syntactic dependencies. The XNLI probe is sourced from the corresponding development split of the XNLI \cite{xnli} dataset. The SPR probing tasks are extracted from the original data by \citet{protoroles}.

\begin{table}
	\centering
	\begin{tabular}{@{}l@{ }rrr@{ }r@{}}
		\toprule
		{} &  tok &  sent &  pred &  arg \\
		\midrule
		CoNLL+SL &   312.2K &       11.3K &        13.3K &        23.9K \\
		SR3de       &    62.6K &        2.8K &         2.9K &         5.5K \\
		\bottomrule
	\end{tabular}
	\caption{Statistics for CoNLL+SemLink (English) and SR3de (German), only core roles.}
	\label{tab:datasets}
\end{table}

\begin{table}[h]
	\centering
	\begin{tabular}{@{}lcrr@{}}
		\toprule
		 &      type & en &     de \\
		\midrule

		*token.ix &  unary & 208.9K &  46.9K \\
		ttype [v]       & unary & 177.2K &  34.0K \\
		lex.unit [v]     & unary & 187.6K &  35.7K \\
		pos          &  unary & 312.2K &  62.6K \\
		deprel       &  binary & 300.9K &  59.8K \\
		role      &  binary & 23.9K &   5.5K \\
		**spr & binary & 9.7K & - \\
		xnli         &  unary &  2.5K &   2.5K \\
		\bottomrule
	\end{tabular}
	\caption{Probing task statistics. Tasks marked with [v] use a most frequent label vocabulary. Here and further, tasks marked with * are regression tasks.}
	\label{tab:taskstats}
\end{table}

\begin{table}
\centering
	\begin{tabular}{@{}lcr@{}}
		\toprule
		language &   en &     de \\
		\midrule
		PropBank      &   5  &   10 \\
		VerbNet      &   23  &   29 \\
		FrameNet      &  189  &  300 \\
		\bottomrule
	\end{tabular}
	\caption{$\#$ of role probe labels by formalism.}
	\label{tab:srlstats}
\end{table}

\subsection{Probing tasks}


Our probing kit spans a wide range of probing tasks, ranging from primitive surface-level tasks mostly utilized as anchors later to high-level semantic tasks that aim to provide a representational upper bound to predicate semantics. We follow the training, test and development splits from the original SR3de, CoNLL-2009 and SPR data. The XNLI task is sourced from the development set and only used for scalar mix analysis. To reduce the number of labels in some of the probing tasks, we collect frequency statistics over the corresponding training sets and only consider up to 250 most frequent labels. Below we define the tasks in order of their complexity, Table \ref{tab:taskstats} provides the probing task statistics, Table \ref{tab:srlstats} compares the categorical role labeling formalisms in terms of granularity, and Table \ref{tab:examples} provides examples. We evaluate the classification performance using Accuracy, while regression tasks are scored via Mean Squared Error.



\begin{table}
	\centering
		\begin{tabular}{@{}lll@{}}
			\toprule
			task & input & label \\
			\midrule
      \texttt{token.ix} &   I [saw] a cat. &$\rightarrow$ 2  \\
			\texttt{ttype} &   I [saw] a cat. &$\rightarrow$ saw  \\
			\texttt{lex.unit} &   I [saw] a cat. &$\rightarrow$ see.V  \\
			\texttt{pos} &   I [saw] a cat. &$\rightarrow$ VBD  \\
			\texttt{deprel} &   [I]\textsubscript{tgt} [saw]\textsubscript{src} a cat. &$\rightarrow$ SBJ  \\
			\texttt{role.vn} &   [I]\textsubscript{tgt} [saw]\textsubscript{src} a cat. &$\rightarrow$ Experiencer  \\
			\texttt{spr.vltn} &   [I]\textsubscript{tgt} [saw]\textsubscript{src} a cat. &$\rightarrow$ 2  \\
			\bottomrule
		\end{tabular}
		\caption{Word-level probing task examples for English. \texttt{vltn} corresponds to the \texttt{volition} SPR property.}%
		\label{tab:examples}
\end{table}

\paragraph{Token type (\texttt{ttype})} predicts the type of a word. This requires contextual processing since a word might consist of several wordpieces;

\noindent\textbf{Token position (\texttt{token.ix})} predicts the linear position of a word, cast as a regression task over the first 20 words in the sentence. Again, the task is non-trivial since it requires the words to be assembled from the wordpieces.

\noindent\textbf{Part-of-speech (\texttt{pos})} predicts the language-specific part-of-speech tag for the given token.

\noindent\textbf{Lexical unit (\texttt{lex.unit})} predicts the lemma and POS of the given word -- a common input representation for the entries in lexical resources. We extract coarse POS tags by using the first character of the language-specific POS tag.

\noindent\textbf{Dependency relation (\texttt{deprel})} predicts the dependency relation between the parent \texttt{src} and dependent \texttt{tgt} tokens;


\noindent\textbf{Semantic role (\texttt{role.[frm]})} predicts the semantic role given a predicate \texttt{src} and an argument \texttt{tgt} token in one of the three role labeling formalisms: PropBank \texttt{pb}, VerbNet \texttt{vn} and FrameNet~\texttt{fn}. Note that we only probe for the role label, and the model has no access to the verb sense information from the data.

\noindent\textbf{Semantic proto-role (\texttt{spr.[prop]})} is a set of eleven regression tasks predicting the values of the proto-role properties as defined in \cite{protoroles}, given a predicate \texttt{src} and an argument \texttt{tgt}.

\noindent\textbf{XNLI} is a sentence-level NLI task directly sourced from the corresponding dataset. Given two sentences, the goal is to determine whether an entailment or a contradiction relationship holds between them. We use NLI to investigate the layer utilization of mBERT for high-level semantic tasks. We extract the sentence pair representation via the \textit{[CLS]} token and treat it as a unary probing task.

\section{Results}

Our models are implemented using AllenNLP.\footnote{We make the code publicly available} We train the probes for 20 epochs using the Adam optimizer with default parameters and a batch size of 32. Due to the frozen encoder and flat model architecture, the total runtime of the experiments is under 8 hours on a single Tesla V100 GPU.

\subsection{General Trends}

\begin{table}
\centering
\begin{tabular}{lrr|lrr}
\toprule
task     &    en   &   de   & task & en & de \\
\midrule
*token.ix &  1.71 &  2.48 & deprel   &  0.95 &  0.95\\
ttype    &  1.00 &  1.00 & role.fn  &  0.92 &  0.59\\
lex.unit &  1.00 &  1.00 & role.pb  &  0.96 &  0.71\\
pos      &  0.97 &  0.97 & role.vn  &  0.94 &  0.73\\

\bottomrule
\end{tabular}
\caption{Best dev score for word-level tasks over 20 epochs ($Acc$ for classification, $MSE$ for regression).}
\label{tab:performance}
\end{table}

While absolute performance is secondary to our analysis, we report the probing task scores on respective development sets in Table \ref{tab:performance}. We observe that grammatical tasks score high, while core role labeling lags behind - in line with the findings of \citet{pipeline}\footnote{Our results are not directly comparable due to the differences in datasets and formalisms.}  We observe lower scores for German role labeling which we attribute to the lack of training data. Surprisingly, as we show below, this doesn't prevent the edge probe from learning to locate relevant role-semantic information in mBERT's layers.

\begin{figure}
  \includegraphics[width=8cm]{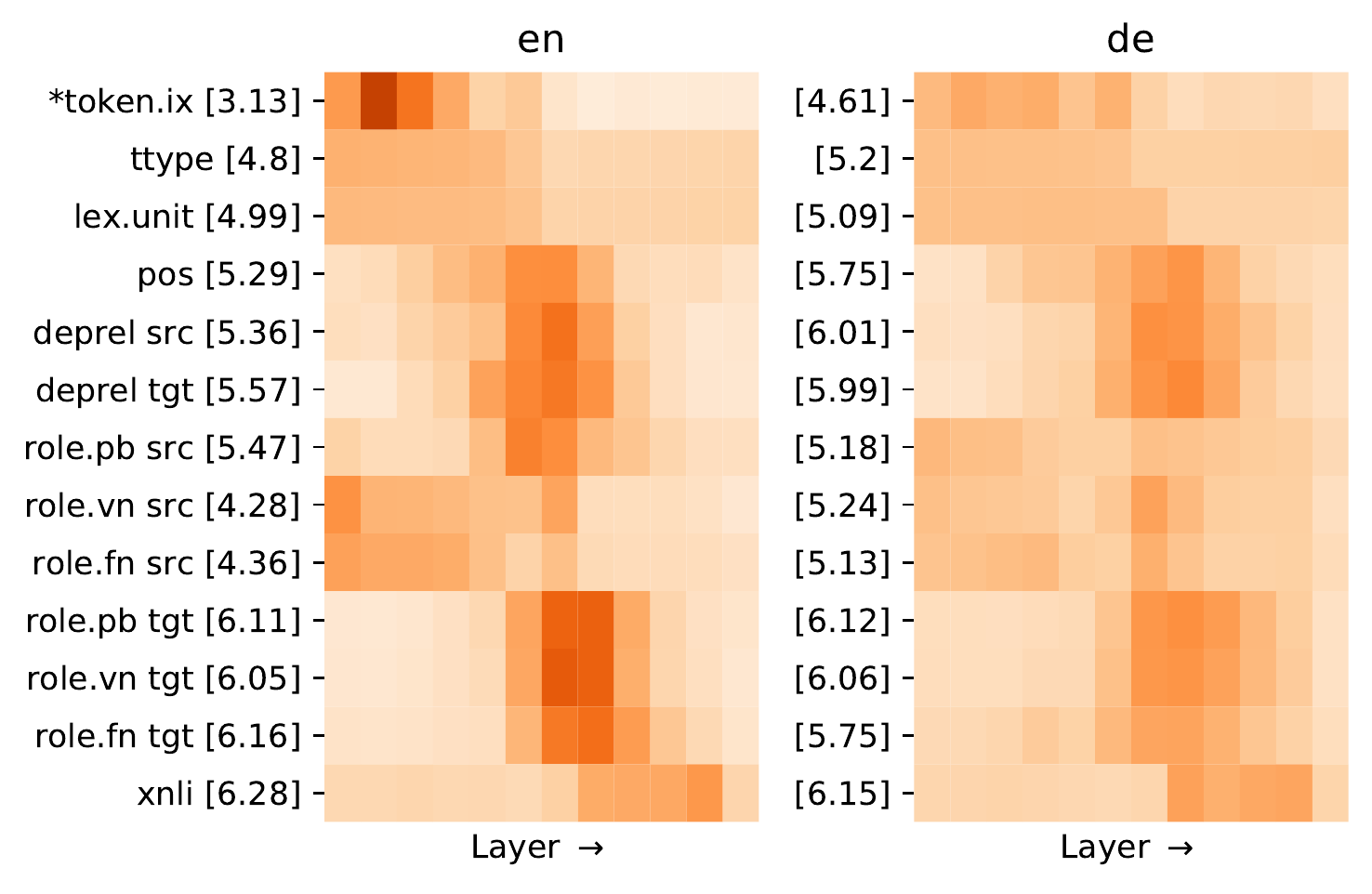}
  \caption{Layer probing results}
  \label{fig:mix_main}
\end{figure}

Our results mirror the findings of \citet{pipeline} about the sequential processing order in BERT. We observe that the layer utilization among tasks (Fig. \ref{fig:mix_main}) aligns for English and German\footnote{Echoing the recent findings on mBERT's multilingual capacity \cite{pires, udify}}, although we note that in terms of center-of-gravity mBERT tends to utilize deeper layers for German probes. Basic word-level tasks are indeed processed early by the model, and XNLI probes focus on deeper levels, suggesting that the representation of higher-level semantic phenomena follows the encoding of syntax and predicate semantics.

\subsection{The Effect of Formalism}

Using separate scalar mixes for source and target tokens allows us to explore the cross-formalism encoding of role semantics by mBERT in detail. Role labeling probe's layer utilization drastically differs for predicate and argument tokens. While the argument representation \texttt{role*tgt} mostly focuses on the same layers as the dependency parsing probe, the layer utilization of the predicates \texttt{role*src} is affected by the chosen formalism. PropBank predicate token mixing weights emphasize the same layers as dependency parsing -- in line with the previously published results. However, the probes for VerbNet and FrameNet predicates (\texttt{role.vn src} and \texttt{role.fn src}) utilize the layers associated with \texttt{ttype} and \texttt{lex.unit} that contain lexical information. Coupled with the fact that both VerbNet and FrameNet assign semantic roles based on lexical-semantic predicate groupings (frames in FrameNet and verb classes in VerbNet), this suggests that the lower layers of mBERT implicitly encode predicate sense information; moreover, sense encoding for VerbNet utilizes deeper layers of the model associated with syntax, in line with VerbNet's predicate classification strategy. This finding confirms that the formalism can indeed have linguistically meaningful effects on probing results.


\subsection{Anchor Tasks in the Pipeline}

\begin{figure}
	\includegraphics[width=8cm]{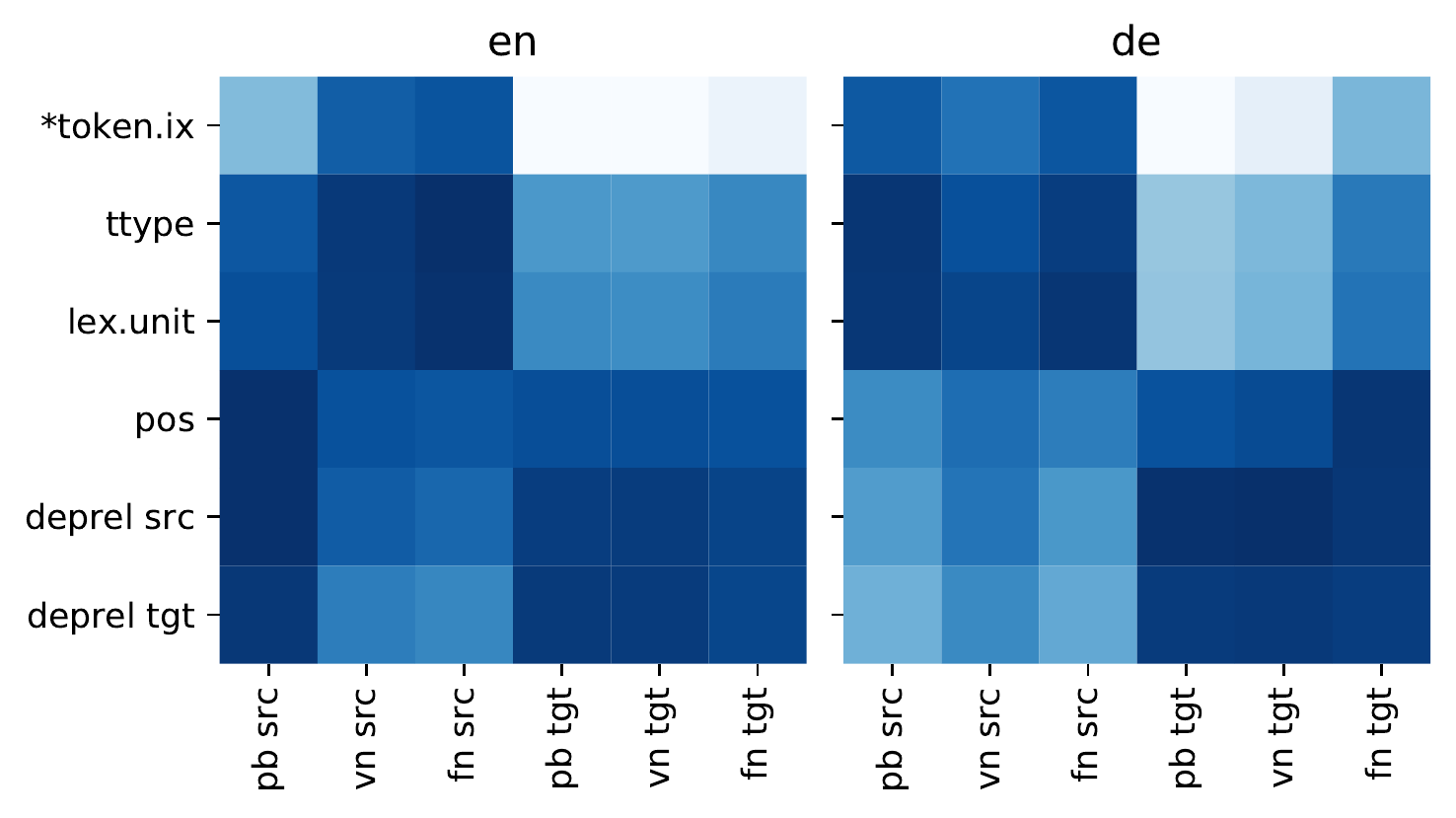}
	\caption{Anchor task analysis of SRL formalisms.}
	\label{fig:mix_main_anchor}
\end{figure}

We now use the scalar mixes of the role labeling probes as target tasks, and lower-level probes as anchor tasks to qualitatively explore the differences between how our role probes learn to represent predicates and semantic arguments\footnote{Darker color corresponds to higher similarity.} (Fig. \ref{fig:mix_main_anchor}). The results reveal a distinctive pattern: while the predicate layer utilization \texttt{src} is similar to the scalar mixes learned for \texttt{ttype} and \texttt{lex.unit}, the learned argument representations \texttt{tgt} attend to the layers associated with dependency relation and POS probes, and the pattern reproduces for English and German. This aligns with the traditional separation of the semantic role labeling task into predicate disambiguation followed by semantic argument identification and labeling, along with the feature sets employed for these tasks \cite{bjorkelund}. Note that the observation about the pipeline-like task processing within the BERT encoders thereby holds, albeit on a sub-task level.

\subsection{Formalism Implementations}

Both layer and anchor task analysis reveal a prominent discrepancy between English and German role probing results: while the PropBank predicate layer utilization for English mostly relies on syntactic information, German PropBank predicates behave similarly to VerbNet and FrameNet. The difference in the number of role labels for English and German PropBank (Table \ref{tab:srlstats}) points at possible qualitative differences in the labeling schemes. The data for English stems from the token-level alignment in SemLink that maps the original PropBank roles to VerbNet and FrameNet. Role annotations for German have a different lineage: they originate from the FrameNet-annotated SALSA corpus \cite{salsa} semi-automatically converted to PropBank style for the CoNLL-2009 shared task \cite{conll2009}, and enriched with VerbNet labels in SR3de \cite{sr3de}. As a result, while English PropBank labels are assigned in a \emph{predicate-independent} manner, German PropBank, following the same numbered labeling scheme, keeps this scheme \emph{consistent within the frame}. We assume that this incentivizes the probe to learn semantic verb groupings and reflects in our probing results. The ability of the probe to detect subtle differences between formalism implementations constitutes a new use case for probing, and a promising direction for future studies.

\subsection{Encoding of Proto-Roles}

We now turn to the probing results for decompositional semantic proto-role labeling tasks. Unlike \cite{edgeprobing} who used a multi-label classification probe, we treat SPR properties as separate regression tasks. The results in Table \ref{tab:performance_spr} show that the performance varies by property, with some of the properties attaining reasonably low MSE scores despite the simplicity of the probe architecture and the small dataset size. We do not observe a clear performance trend depending on whether the property is associated with Proto-Agent or Patient.

\begin{table}
\centering
\begin{tabular}{lr}
\toprule
property &  MSE \\
\midrule
(A) *instigation        &       1.12 \\
(A) *volition           &       0.87 \\
(A) *awareness          &       0.79 \\
(A) *sentient           &       0.47 \\
(A) *change.of.location &       0.66 \\
(A) *exists.as.physical &       1.29 \\
\midrule
(P) *created            &       0.87 \\
(P) *destroyed          &       0.61 \\
(P) *changes.possession &       0.83 \\
(P) *change.of.state    &       1.24 \\
(P) *stationary         &       0.77 \\
\bottomrule
\end{tabular}
\caption{Best dev MSE for proto-role probing tasks over 20 epochs. A - Proto-Agent, P - Proto-Patient.}
\label{tab:performance_spr}
\end{table}

\begin{figure}
	\includegraphics[width=8cm]{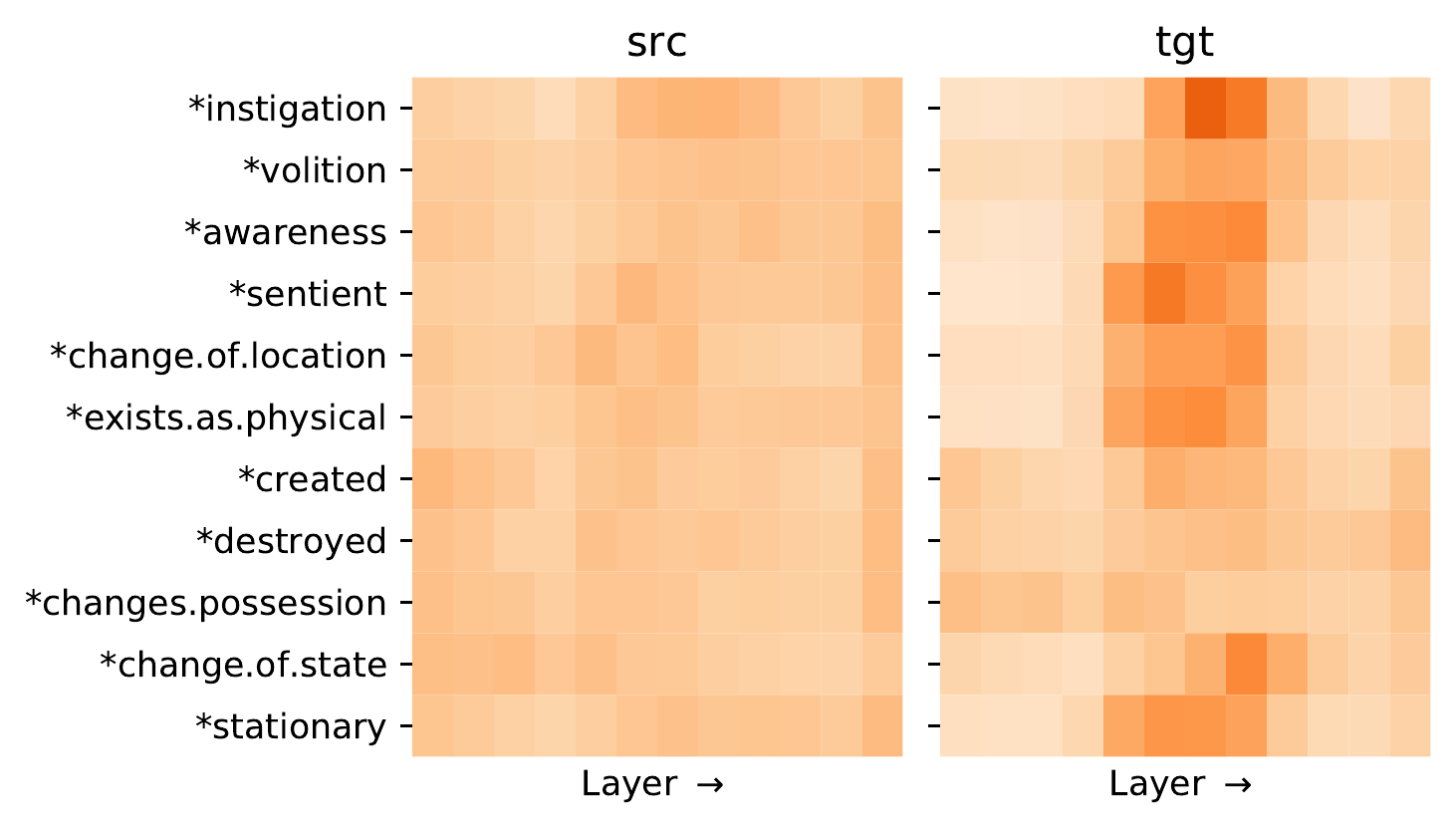}
	\caption{Layer utilization for SPR properties.}
	\label{fig:mix_spr}
\end{figure}

Our fine-grained, property-level task design allows for more detailed insights into the layer utilization by the SPR probes (Fig. \ref{fig:mix_spr}). The results indicate that while the layer utilization on the predicate side (\texttt{src}) shows no clear preference for particular layers (similar to the results obtained by \citet{pipeline}), some of the proto-role features follow the pattern seen in the categorical role labeling and dependency parsing tasks for the argument tokens \texttt{tgt}. With few exceptions, we observe that the properties displaying that behavior are Proto-Agent properties; moreover, a close examination of the results on syntactic preference by \citet[p. 483]{protoroles} reveals that these properties are also the ones with strong preference for the subject position, including the outlier case of \texttt{stationery} which in their data behaves like a Proto-Agent property. The correspondence is not strict, and we leave an in-depth investigation of the reasons behind these discrepancies for follow-up work.

\section{Conclusion}

We have demonstrated that the choice of linguistic formalism can have substantial, linguistically meaningful effects on role-semantic probing results. We have shown how probing classifiers can be used to detect discrepancies between formalism implementations, and presented evidence of semantic proto-role encoding in the pre-trained mBERT model. Our refined implementation of the edge probing framework coupled with the anchor task methodology enabled new insights into the processing of predicate-semantic information within mBERT. Our findings show that linguistic formalism is an important factor to be accounted for in probing studies. While our work illustrates this point using a single task and a single probing framework, the influence of linguistic formalism per se is likely to be present for any probing setup that builds upon linguistic material. An investigation of how, whether, and why formalisms affect probing results for tasks beyond role labeling and for frameworks beyond edge probing constitutes an exciting avenue for future research.

\bibliography{bib}
\bibliographystyle{acl_natbib}





\end{document}